# Chinese Poetry Generation with Flexible Styles

*Jiyuan Zhang*[1], *Dong Wang*[1]

[1]Center for Speech and Language Technologies, Tsinghua University
wangdong99@mails.tsinghua.edu.cn

## Abstract

Research has shown that sequence-to-sequence neural models, particularly those with the attention mechanism, can successfully generate classical Chinese poems. However, neural models are not capable of generating poems that match specific styles, such as the impulsive style of Li Bai, a famous poet in the Tang Dynasty. This work proposes a memory augmented neural model to enable the generation of style-specific poetry. The key idea is a memory structure that stores how poems with a desired style were generated by humans, and uses similar fragments to adjust the generation. We demonstrate that the proposed algorithm generates poems with flexible styles, including styles of a particular era and an individual poet.

## 1. Introduction

Classical Chinese poems have been a source of fascination for thousands of years, covering many different Chinese dynasties. Various genres have flourished at different points in history, with the most typical examples including Tang poetry, Song iambics, Ming poetry and Qing poetry. In this paper, we are particularly interested in machine generated poetry. There are two motivations behind this study. Firstly, we would like to demonstrate that the writing of poetry, although an artistic activity, is largely an empirical process, which given an adequate quantity of data to learn from, can therefore pragmatically be imitated by machines. Secondly, systems like those proposed in this paper could be used to assist human poets by providing guidance, drafts, or suggesting some particularly innovative expressions that could improve the work of humans.

In Chinese poetry, different genres have different regulations regarding structure and pronunciation in order for a poem to be considered part of a particular genre. This paper focuses on one specific genre, quatrains. Poems in the quatrain genre follow a strict structure (four lines with five or seven characters per line), a regulated rhythmical form (the last characters in the second and fourth lines must follow the same rhythm), and a required tonal pattern (tones of characters in some positions should satisfy a predefined regulation) [1]. An example of a quatrain written by Wei Wang, a famous poet in the Tang Dynasty, is shown in Table 1. The translation is from [2].

There are a number of different approaches that have been developed for generating Chinese poetry. Several approaches have been studied by researchers for Chinese poem generation. These include probabilistic approaches, which create various probabilistic models in order to represent the temporal dynamics of words [3, 4], symbolic approaches, which search a large corpus in order to find suitable sentences, and then recompose the retrieved sentences [5], and neural approaches, which involves the creation of various forms of neural models that project the meaning of words in a continuous concept space (via word embedding) where the temporal dynamics are learned [6, 7, 8].

The key difference with these approaches is that symbolic and probabilistic approaches focus on the surface form of words, i.e. the individual characters, whereas neural approaches are based on the inherent meaning of words and sentences, leading to deeper understanding of poems and the generation process. This difference is particularly crucial for learning Chinese poems, because in Chinese poetry, many poems have a very similar deeper meaning (such as the focus on a particular topic), but with wildly different surface forms. This means that two poems with a very similar meaning can be written in a completely different way, and therefore means that the underlying meaning is potentially easier to model. Several authors have reported that neural approaches, with various model structures, can generate poems that are regarded to be both more fluent and more meaningful than competing non-neural methods [7, 9, 8].

Table 1: An example of a 5-char quatrain. The tonal pattern is shown the end of each line, where 'P' indicate a level tone, 'Z' indicates a downward tone, and '*' indicates the tone can be either.

| 相思 |
|---|
| Yearning |
| 红豆生南国，（P Z P P P） |
| These red beans grow only in the south, |
| 春来发几枝。（P P P Z P） |
| In spring the branch away and flourish. |
| 愿君多采撷，（Z P P Z P） |
| Why don't you harvest plentiful to cherish, |
| 此物最相思。（Z Z Z P P） |
| For they are most symbolishing of yearning. |

However, one limitation with many existing neural-based algorithms is that despite promising results, they are less capable of fine-grained poetry generation control, e.g., it is difficult to generate poems that correspond to a desirable style. In this paper, when we use the term 'style', we refer to the way that poetry is expressed, which includes word selection and composition. This limitation is due to the nature of neural models, which as mentioned previously, build a temporal model based on deeper meaning, rather than focusing on the surface form (i.e. word choice). However, the style of a poem is defined by how the surface forms are chosen and used, rather than the underlying meaning. This implies that neural models tend to generate fluent sentences by selecting the most frequently used words (in the model's training set) that can express the meaning, but do not care much about the style of expression. This 'trivial' generation is not our goal, our aim is for machines to be able to generate more interesting poems which exhibit a specific desirable style. One possible solution is to train style-dependent models, but this is not feasible for styles with limited training data, e.g., a style of a preferred poet. Model adaptation is possible, but adapting neural models is difficult, partly due to the fact th at the model parameters are heavily shared, and most importantly, our approach aims to be capable of quick and flexible style generation, that can deal with subtle styles defined by users, without the need for model retraining. It should also be observed that while the probabilistic and symbolic approaches are focused on the surface forms, and thus the style, the poetry produced is less fluent and meaningful, and so this is not a suitable solution to this problem.

This paper presents a memory-augmented neural model to solve the problem of quickly and flexibly generating Chinese poems that correspond a very specific style. To do this, we augment a conventional neural generation model with a memory structure which is specially constructed and contains only relevant sample poems that correspond to a desired style. Moreover, the memory is constructed depending on the users' input, leading to a local memory approach with a more accurate style specification. The model output is a combination of both the generation from the neural model and also from this local memory. This combination aims to encourage the same word usage as the poems in the memory, leading to a generation that corresponds to the desired style but is still fluent and meaningful. In this sense, the memory can be regarded as a flexible regularization that can guide the generation of any style, on condition that a number of some samples of that style of poetry can be obtained. To demonstrate the success of our method, evaluate with human poetry experts confirmed that this new combined approach can generate poems that correspond to both the styles of specific poets, and also the stles corresponding to specific eras.

## 2. Related Work

Symbolic poem generation involves various search methods on poem fragments [10, 11, 12, 13, 14, 5], and the probabilistic approach involves various statistical machine translation (SMT) methods [3, 4].

More recently, neural models have been the subject of much attention due to them being able to more deeply understand the meaning of a poem (as discussed previously). In this section, we only review Chinese poetry generation methods that are directly relevant to the our research. One relevant study is by [6], which proposed an RNN-based approach that produces each new line of poetry character-by-character using a recurrent neural network(RNN), with all the lines generated already (in the form of a vector) as a contextual input. This model can generate quatrains of reasonable quality. [7] proposed a much simpler neural model that treats a poem as an entire character sequence, and poem generation is therefore conducted character-by-character. This approach is easy to extend to other various genres such as Song Iambics. To avoid theme drift caused by this longsequence generation, [7] utilized the neural attention mechanism[15] by which human intention is encoded by an RNN in order to guide the generation. The same approach was utilized by [9] for Chinese quatrain generation. [8] also proposed a similar sequence generation model, but when each character is generated, attention is also focused on all characters that have been generated in the sequence as well as the human input. They also proposed a topic planning scheme to encourages a smooth and consistent theme. Recently, [16] developed hierarchical RNN model that conducts iterative generation, using the output of the last iteration as an input of the next iteration for refinement. However, it should be noted that none of these existing neural methods solved the issue of being able to generate poetry in specific styles, and being able to change between styles without retraining the model.

The proposed memory argumentation approach was inspired by the recent advance in the neural Turing machine [17, 18] and memory network [19]. These new models equip neural networks with an external memory that can be accessed and manipulated via some trainable operations. In comparison to these, the memory augmentation in this research simply fulfills the role of knowledge storage, and the only operation that needs to be carried out is a simple pre-defined READ. In this sense, our model can be regarded as a simplified neural Turing machine that omits training. Preliminary experiments found that this simple model performs better than more complex models that are quipped with flexible operations and elaborate training, probably due to poems being unique, and there being limited examples of repeated training data.

## 3. Local Memory augmentation

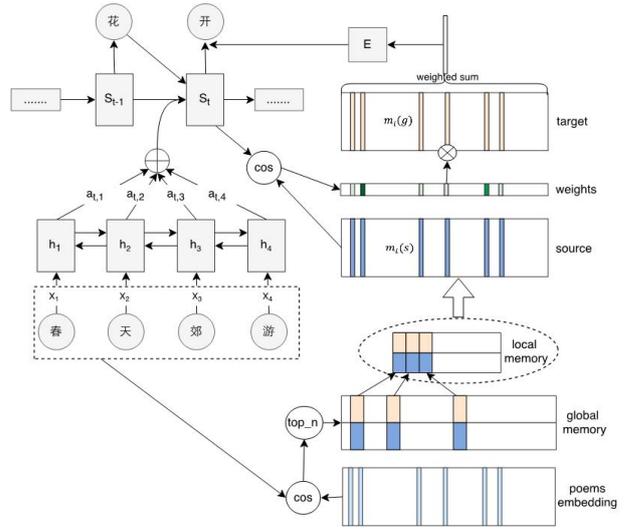

Figure 1: *The memory-augmented neural model used for Chinese poetry generation..*

Fig. 1 shows our proposed approach, featuring two components: a neural model (left) and a local memory (right). The neural model is an attention-based sequence-to-sequence model first proposed by [15] and utilized for Chinese poem generation by [7]. The encoder is a bi-directional RNN with GRU units and is used to convert the input topic words, denoted by the embeddings of the compositional characters $(x_1, x_2, \cdots, x_N)$, into a sequence of hidden states $(h_1, h_2, \cdots, h_N)$. The decoder then generates the whole quatrain character-by-character, denoted by the corresponding embeddings $(y_1, y_2, \cdots, y_N)$. At each step t, the state of the decoder is updated as follows:

$$s_t = f_d\left(y_{t-1}, s_{t-1}, \sum_{t=1}^{N} \alpha_{t,i} h_i\right)$$

Where $\alpha_{t,i}$ represents the 'attention' on $h_i$ at the present generation,and can be implemented as any function in the form $\alpha_{t,i} = g\left(s_{t-1}, h_i, \{h_j\}\right)$. The model output is a posterior probability over the whole character set, where W is the projection parameter, given as:

$$z_t = \sigma(s_t W)$$

The memory is a set of elements $\{m_i\}_{i=1}^{K}$, of size K. Each element $m_i$ has two parts, the source, $m_i(s)$, that encodes the context, i.e. when this element should be selected, and the target, $m_i(g)$, that encodes the output within that context. To construct the memory, we firstly train the neural model, and then produce the memory by running the neural model decoder with some selected sample poems. For the k-th selected poem, the character sequence is input to the decoder one-by-one, with the contribution from the encoder set to zero. Let $p_k$ denote the starting position of this poem in the memory, the decoder state at the j-th step is used as the source part of the $(p_k + j)$-th memory element, and the corresponding character embedding, $x_j$, is set to be the target. This is defined as:

$$m_i(s) = f_d(x_{j-1}, s_{j-1}, 0)$$
and
$$m_i(g) = x_j; i = p_k + j. \qquad (1)$$

At run-time, the memory output is computed by averaging the targets of all the elements, weighted by the distance between the state of the current decoding and the source of each element, written as:

$$v_t = \sum_{i=1}^{K} \cos(s_{t-1}, m_i(s)) m_i(g) \qquad (2)$$

The final output is a linear combination of both the neural model and memory outputs:

$$z_t = \sigma(s_t W + \beta v_t E) \qquad (3)$$

where $\beta$ is a pre-defined weighting factor, and E contains word embeddings of all characters.

A potential problem is that using a lot of sample poems would result a large memory, and most of the elements are irrelevant to users' input. We propose a local memory approach. As shown in Fig 1 (bottom-right), the top-n sample poems most relevant to user input are selected, according to the cosine distance of the averaged word vectors of the user's input and each poem in the global memory. The corresponding memory elements of the selected poems are collected, forming a local memory that relevant to the present generation task. This reduces memory usage and computational requirements, and results in more theme consistency as all memory elements are relevant to the user's input.

Only the neural model needs to be trained, using sentence-based training. For each line of poetry in the training dataset, a number of characters are sampled to become the input sequence, with the full line being the output. The cross entropy between the distributions over Chinese characters given by the decoder and the ground truth is used as the objective function. The optimization uses the SGD algorithm along with AdaDelta for learning rate [20] adjustment.

## 4. Style Specific Memory

Chinese poems from different eras generally exhibit noticeably different styles. For example, in the Han and Wei Dynasties (the 2nd-3rd century AD), poems (especially quatrains) were derived from folk songs, so the restricts on rhythm and tone are not very stringent. This was changed in Qi and Liang Dynasties (5th-6th century AD), where the social position of poets was largely improved, and many royal poets created a very different style with more constrains, featuring amorous words, elegant expressions and musical pronunciations. This was fundamentally changed in the Tang and Song Dynasties (8th-9th century AD), where more diverse poems were created, combining musical patterns and real life feelings.

There are also style discrepancies between poets, such as between three famous poets, Li Bai, Bai Juyi, and Jia Dao. Li Bai specialized in describing large-scale scenarios and his works are full of magical imagination, whereas Bai Juyi tended to describe homely subjects, and the style of Jia Dao is tranquil, favored deep insight behind normal observations.

The local memory approach, which constructs an input-dependent memory for each generation, is particularly valuable for styled-poem generation. For example, to generate Li Bai's style, using his entire library of more than 800 poems would cause style mixing as the poems focus on diverse topics. Our local memory approach selects only the most relevant poems that match user input, leading to a more accurate topic-dependent style. For this reason, all of our experiments use a local memory.

## 5. Experiments

### 5.1. Configuration

We used the attention-based RNN system by [9] as the baseline. This system was shown to be flexible, powerful, and capable of generating different genres of Chinese poems (and able to fool human experts in many cases). The baseline system was trained with a quatrain database that contains 9,000 5-char and 49,000 7-char quatrains. A larger database which contains 284,899 traditional Chinese poems from a range of genres was used to learn the word embeddings. We tested on generation for three era styles: Han-Wei style (HW), Qi-Liang style (QL) and Song style (SONG) using 100 poems from each era to construct the memory, and three poet styles: Li Bai (LB), Jia Dao (JD) and Bai Juyi (BJY), selecting 200 poems from each for memory creation.

Evaluation is highly challenging, requiring experts with advanced history, culture and literature knowledge, and familiarity with all the styles being tested. We collaborated with many poem generation and criticism experts, most of coming from the Chinese Academy of Social Science (CASS) and Peking University. We asked the experts to recommend each other for the evaluation task, resulting in 5 experts being selected to participate.

### 5.2. Datasets

The baseline system used two customized datasets. The first dataset is a Chinese poem corpus (CPC), which we created and was utilized to train the Chinese character embeddings. This contains 284,899 traditional Chinese poems from a range of genres, including Tang quatrains, Yuan Songs, Song Iambics, and Ming and Qing poems. This large dataset ensures reliable learning for the semantic content of the majority of Chinese characters. The second dataset was also created by us and is a Chinese quatrain corpus (CQC) collected from the internet, containing 9,195 5-char quatrains and 49,162 7-char quatrains. 9,000 5-char and 49,000 7-char quatrains were used for training the at tention model, and the rest for validation.

The third database was used to generate memories with different styles. The first part contains three subsets, each consisting of over 100 poems and containing only poems from one particular era, and the second part contains three subsets, each consisting of about 200 poems from an individual poet. There is some overlap between this database and CQC, but some poems were newly collected to represent particulary styles, as suggested by some poetry professionals. The datasets are summarized in Table 2.

Table 1: *Datasets used in the expriment.*

| Dataset | No. of Poems | Purpose |
|---------|--------------|---------|
| CPC | 284,899 | For word embedding |
| CQC | 58,000 | For neural models |
| Mem-I-Wei | 111 | For Wei Dynasty |
| Mem-I-Tang | 171 | For Qi Dynasty |
| Mem-I-Song | 200 | For Song Dynasty |
| Mem-II-LB | 678 | For Li Bai |
| Mem-II-JD | 316 | For Jia Dao |
| Mem-II-BJY | 1269 | For Bai juyi |

### 5.3. Era sytle

In this experiment, we used 160 topic keywords selected from Shixuhanyinge [21], and experimented with three era styles: Han-Wei (HW), Qi-Liang (QL) and Song (SONG). Expert consensus was that era styles are identifiable only by considering multiple poems, and we therefore designed a group-based evaluation: groups of poems were presented to the experts, with each containing 5 poems of the same style, and the experts were asked to identify the era style. We compared this with poems generated with the baseline, which was assumed to generate an 'unclear' style. We collected 48 evaluation samples (groups) in total.

The results in Table 3 show that for the Han-Wei and Qi-Liang styles, the experts could successfully and consistently identify the styles (probability of 0.92), however the Song style was easily confused with the baseline system. Analysis showed that this is because the data used to train the baseline system are mostly from the Tang and Song Dynasties, with both styles being very similar due to their proximity in history. Therefore the baseline system is biased towards the Song style.

Table 3: *Probability that poems generated by each configuration were labeled as various era styles..*

|  | Probability | | | |
|---|---|---|---|---|
| Model | HW | QL | SONG | Unclear |
| Baseline | 0.08 | 0.17 | 0.41 | 0.34 |
| +HW Mem | 0.92 | 0.0 | 0.0 | 0.08 |
| +QL Mem | 0.00 | 0.92 | 0.0 | 0.08 |
| +SONG Mem | 0.08 | 0.08 | 0.50 | 0.34 |

### 5.4. Poet sytle

In this experiment, we used 160 topic keywords selected from Shixuhanyinge [21], and experimented with three era styles: Similar experiments were conducted for poet style, with the difference that poet topics were chosen based on the words that mostly appear in the poet's poems, as the topic was identified by experts as being a distinct indicator of a poet.

We collected 48 evaluation samples in total. The evaluation results are shown in Table 5, and show that the three styles can be distinctly identified, with 50% of the poems being correctly identified in all cases. However, experts tend to classify a large proportion of poems as being an 'unclear' style. Poetry experts assisted with the analysis of the results, and theorize that most experts tend to recognize styles by 'indicating objects', such as mountain or river names, official positions, temples, or monks. However, all the poets selected were from the Tang Dynasty, so the indicating objects are similar. This caused some confusion for the experts, leading to incorrect identifications. Nevertheless, some highly knowledgeable experts identified all the styles correctly, demonstrating that the styles had indeed been learned. However these are arguably too subtle to be noticed by the experts without a detailed knowledge of these exact poets.

Another interesting observation is that many poems generated by the baseline were identified as being LB and BJY style, but few were identified as JD style. One reason might be the bias due to the proportion of these poets in the training data: the numbers of poems for the three poets are 678(LB), 316(JD), and 1269(BJY), respectively. It is also possible that the experts are more familiar with LB and BJY and so tend to select them with more confidence.

Table 4: *Probability that poems generated by each configuration were labeled as various poet styles..*

|  | Probability | | | |
|---|---|---|---|---|
| Model | LB | JD | BJY | Unclear |
| Baseline | 0.34 | 0.08 | 0.25 | 0.33 |
| +LB Mem | 0.50 | 0.16 | 0.00 | 0.34 |
| +JD Mem | 0.16 | 0.50 | 0.16 | 0.18 |
| +BJY Mem | 0.08 | 0.17 | 0.50 | 0.25 |

Table 5: *A quatrain example generated by the memory approach imitating LiBai style.*

| 山亭 |
|---|
| Mountain Kiosk |
| 西望山城上， |
| To gaze west from the mountain, |
| 高峰眺汉亭。 |
| A kiosk in the peak far away. |
| 水云如雪碧， |
| Cloud and brook all are green, |
| 石远下江青。 |
| Cyan crock far away in the water. |

## 6. Conclusions

This paper proposed a memory-augmented neural model to generate poems with distinct styles. The key contribution was to regularize the model with a memory structure containing a set of poems of a chosen style, encouraging generation of poems of that style. Our method can generate poems that can imitate styles of different eras and distinct poets. Future work would involve investigating more effective approaches for combining the model and memory components, and also considering the use of different sparse memory methods.